\documentclass[conference]{IEEEtran}
\IEEEoverridecommandlockouts

\usepackage{cite}
\usepackage{amsmath,amssymb,amsfonts}
\usepackage{amsthm}
\usepackage{algorithmic}
\usepackage{algorithm}
\usepackage{graphicx}
\usepackage{mathtools}
\usepackage{textcomp}
\usepackage{xcolor}
\usepackage{multirow}
\usepackage{dsfont}
\usepackage{booktabs}
\usepackage{dsfont}
\usepackage{makecell}
\usepackage{balance}
\usepackage[nolist]{acronym}
\usepackage{tikz}
\usepackage{svg}
\usepackage{subcaption}
\usetikzlibrary{positioning,arrows.meta,decorations.pathreplacing,calc,fit}

\newtheorem{definition}{Definition}

\begin{acronym}[XXX] 
\acro{ai}[AI]{Artificial Intelligence}
\acro{auc}[AUC]{Area Under the Curve}
\acro{bva}[BVA]{Best Validation Accuracy}
\acro{cnn}[CNN]{Convolutional Neural Network}
\acro{dl}[DL]{Deep Learning}
\acro{dnn}[DNN]{Deep Neural Network}
\acro{fl}[FL]{Federated Learning}
\acro{ff}[FF]{Feedforward Neural Network}
\acro{gelu}[GELU]{Gaussian Error Linear Unit}
\acro{glai}[GLAI]{GreenLightningAI}
\acro{kd}[KD]{Knowledge Distillation}
\acro{ml}[ML]{Machine Learning}
\acro{mlp}[MLP]{Multilayer Perceptron}
\acro{mva}[MVA]{Maximum Validation Accuracy}
\acro{mha}[MHA]{Multihead Attention}
\acro{moe}[MoE]{Mixture-of-Experts}
\acro{mimick}[NM]{Neural Mimicking}
\acro{nlp}[NLP]{Natural Language Processing}
\acro{cv}[CV]{Computer Vision}
\acro{nn}[NN]{Neural Network}
\acro{oui}[OUI]{Overfitting-Underfitting Indicator}
\acro{prelu}[PReLU]{Parametric Rectified Linear Unit}
\acro{poc}[PoC]{Proof-of-Concept}
\acro{qk}[QK]{Quantitative Knowledge}
\acro{relu}[ReLU]{Rectified Linear Unit}
\acro{silu}[SiLU]{Sigmoid Linear Unit}
\acro{sgd}[SGD]{Stochastic Gradient Descent}
\acro{svd}[SVD]{Singular Value Decomposition}
\acro{sota}[SOTA]{state-of-the-art}
\acro{sk}[SK]{Structural Knowledge}
\acro{tl}[TL]{Transfer Learning}
\acro{vt}[VT]{Vision Transformer}
\acro{vit}[ViT]{Vision Transformer}
\acro{wd}[WD]{Weight Decay}
\end{acronym}
\begin{document}

\title{\textsc{FedSQ}:\\ Optimized Weight Averaging via Fixed Gating
}

\author{
\makebox[.33\linewidth][c]{
\begin{tabular}{c}
Cristian Pérez-Corral*\thanks{*Corresponding author} \\
\textit{Universitat Politècnica de València} \\
Valencia, Spain \\
cpercor@upv.es
\end{tabular}
}
\makebox[.33\linewidth][c]{
\begin{tabular}{c}
Jose I. Mestre \\
\textit{Universitat Politècnica de València} \\
Valencia, Spain \\
jimesmir@disca.upv.es
\end{tabular}
}

\makebox[.33\linewidth][c]{
\begin{tabular}{c}
Alberto Fernández-Hernández \\
\textit{Universitat Politècnica de València} \\
Valencia, Spain \\
a.fernandez@upv.es
\end{tabular}
}\\[1em]
\makebox[.33\linewidth][c]{
\begin{tabular}{c}
Manuel F. Dolz \\
\textit{Universitat Jaume I} \\
Castelló de la Plana, Spain \\
dolzm@uji.es
\end{tabular}
}
\makebox[.33\linewidth][c]{
\begin{tabular}{c}
José Duato \\
\textit{Openchip \& Software Technologies} \\
Barcelona, Spain \\
jose.duato@openchip.es
\end{tabular}
}
\makebox[.33\linewidth][c]{
\begin{tabular}{c}
Enrique S. Quintana-Ortí \\
\textit{Universitat Politècnica de València} \\
Valencia, Spain \\
quintana@disca.upv.es
\end{tabular}
}
}

\maketitle

\begin{abstract}
Federated learning (FL) enables collaborative training across organizations without sharing raw data, but it is hindered by statistical heterogeneity (non-i.i.d.\ client data) and by instability of naive weight averaging under client drift. In many cross-silo deployments, FL is warm-started from a strong pretrained backbone (e.g., ImageNet-1K) and then adapted to local domains. Motivated by recent evidence that ReLU-like gating regimes (structural knowledge) stabilize earlier than the remaining parameter values (quantitative knowledge), we propose \textsc{FedSQ} (Federated Structural-Quantitative learning), a transfer-initialized neural federated procedure based on a DualCopy, piecewise-linear view of deep networks. \textsc{FedSQ} freezes a structural copy of the pretrained model to induce fixed binary gating masks during federated fine-tuning, while only a quantitative copy is optimized locally and aggregated across rounds. Fixing the gating reduces learning to within-regime affine refinements, which stabilizes aggregation under heterogeneous partitions. Experiments on two convolutional neural network backbones under i.i.d.\ and Dirichlet splits show that \textsc{FedSQ} improves robustness and can reduce rounds-to-best validation performance relative to standard baselines while preserving accuracy in the transfer setting.
\end{abstract}

\section{Introduction}
The proliferation of connected devices and the digitization of organizational processes have led to an unprecedented amount of data being generated in distributed settings. In many real-world scenarios, collecting such data in a centralized repository is impractical or undesirable due to privacy, legal, or operational constraints. In this context, \ac{fl} has emerged as a paradigm to collaboratively train \ac{ml} and \ac{dl} models while keeping raw data local to each participant~\cite{mcmahan2017,kairouz2019}.

Despite its appeal, \acs{fl} introduces fundamental challenges that are absent or less pronounced in centralized training. First, client data is often statistically heterogeneous (non-i.i.d.), which can lead to unstable or slow convergence and to suboptimal global models when using standard federated averaging~\cite{mcmahan2017}. This heterogeneity may stem from label distribution shifts, feature shifts, or domain shifts across clients, motivating a large body of work aimed at improving robustness under non-i.i.d. data. Representative examples include approaches based on local normalization to address feature shift (\textsc{FedBN})~\cite{xiaoxiao2021}, methods that exploit shared representations with personalized local components (\textsc{FedRep})~\cite{collins2021}, and architectures that explicitly account for heterogeneity through specialization, such as federated mixtures of experts (\textsc{FedMix})~\cite{reisser2021}.

Second, communication between server and clients frecuently becomes a dominant bottleneck, particularly when models are large or bandwidth is limited. Beyond reducing the number of communication rounds by increasing local computation (as in federated averaging), a complementary line of work targets communication efficiency through structured updates~\cite{konecny2016}, gradient quantization~\cite{alistarh2017}, aggressive sparsification~\cite{lin2018}, or compression schemes tailored to federated non-i.i.d. regimes (e.g., STC)~\cite{sattler2019}. These methods highlight a central trade-off in \acs{fl}: improving statistical efficiency (accuracy and convergence) while minimizing system costs (communication, latency, and client-side computation).

A practical aspect often overlooked in methodological discussions is that many real deployments of \acs{fl} do not start from random initialization. Instead, it is common to warm-start federated training from a backbone pre-trained on large-scale public data (e.g., ImageNet-1K in computer vision) and then perform federated fine-tuning on the private client datasets. Recent systematic studies show that such pre-trained initialization can reduce the number of communication rounds required to reach a target accuracy and can dampen the adverse effects of data and system heterogeneity in federated optimization~\cite{nguyen2022, chen2023on, zhou2025}. This practice is also consistent with classical \ac{tl} evidence that early representations learned on large visual corpora tend to be broadly reusable across downstream tasks~\cite{yosinski2014transferable}.

Recent work suggests that learning dynamics in ReLU-based \acp{dnn} can be fruitfully understood through two complementary components:~\ac{sk} and \ac{qk}. \ac{sk} corresponds to the gating behaviour induced by ReLU nonlinearities, i.e., the activation masks (activation patterns) that select active paths and thus, the piecewise-affine region in which the network operates for a given input. \ac{qk} corresponds to the numerical values of weights and biases that determine the affine mapping within those active regions. Empirical evidence~\cite{perezcorral2026} suggests that activation-pattern changes decay substantially earlier than weight-update magnitudes across multiple architectures, motivating a two-phase interpretation of training. Along the same lines, Duato et al.~\cite{duato2025} formalize the \ac{sk}/\ac{qk} separation and demonstrate that freezing the structural component while retraining only \ac{qk} can accelerate (re-)training, with direct implications for incremental and federated re-training.

Inspired by these findings, we propose \textsc{FedSQ}, where all participants start from the same ImageNet-1K pretrained initialization, which acts as the fixed structural component, while clients communicate and aggregate only \ac{qk} updates. This induces a late-stage specialization in which optimization proceeds largely within stable activation regimes, reducing client-side compute and improving performance per round in the federated stage.

Therefore, in this paper we make the following contributions:
\begin{itemize}
    \item We introduce \textsc{FedSQ}, a transfer-initialized two-stage \acs{fl} procedure that decouples \emph{structural} (gating) from \emph{quantitative} (weights) knowledge for cross-silo settings.
    \item We design a federated protocol that freezes the structural component and aggregates only quantitative updates.
    \item We empirically evaluate \textsc{FedSQ} against standard federated baselines under heterogeneous client data and report improvements in convergence efficiency.
\end{itemize}

\section{Related Work}
\ac{fl} trains models across distributed data holders without moving raw data, typically via iterative rounds of local optimization followed by server-side aggregation~\cite{mcmahan2017}. In practical deployments, cross-silo \ac{fl} involves a small number of stable clients with reliable connectivity, while cross-device \ac{fl} targets massive numbers of intermittently available devices; these regimes differ in their dominant bottlenecks~\cite{kairouz2019}. A core difficulty is that client data are typically non-i.i.d., yielding slow or unstable convergence under vanilla \textsc{FedAvg} due to \emph{client drift}. \textsc{FedProx} mitigates client drift with a proximal term discouraging excessive deviation from the global iterate~\cite{li2020}. Adaptive server-side optimization (\textsc{FedOpt}, including \textsc{FedAdam}/\textsc{FedYogi}) improves robustness and tuning under heterogeneous data~\cite{reddi2021}. Beyond label skew, non-i.i.d.\ feature distributions can cause mismatched batch-normalization statistics and degrade aggregation quality; \textsc{FedBN} addresses this by keeping batch normalization statistics local~\cite{xiaoxiao2021}. Another branch targets personalization by learning shared representations with client-specific components, such as \textsc{FedRep}~\cite{collins2021} or \textsc{FedBABU}~\cite{oh2022}. 

\ac{tl} is the de facto starting point in many cross-silo vision deployments: models are initialized from pretrained backbones and then adapted to downstream domains. The transferability of deep representations has been extensively studied; notably, some studies~\cite{yosinski2014transferable} quantify how transferability decreases from early to late layers and identify co-adaptation as a practical obstacle. These motivates our design choice to \textit{(i)} initialize from a strong pretrained model and \textit{(ii)} constrain the subsequent federated adaptation to the subset of parameters that most benefit from domain-specific refinement, which appears to mitigates the umbalanced and drift effect.

Our method relates to an emerging perspective that many \acp{dnn} can be viewed through a \emph{gating} lens, where a discrete routing/selection mechanism determines which computational substructures are active, and the remaining parameters refine the mapping within those selected regimes. While the original \ac{sk}/\ac{qk} decomposition in~\cite{perezcorral2026,duato2025} is articulated for ReLU architectures (where the gate is naturally binary), \textsc{FedSQ} does not assume ReLU nor any specific activation in the pretrained backbone. Instead, we explicitly introduce a \emph{binary} gating surrogate, enforcing a hard (ReLU-like) activation mask during the federated stage, regardless of the activation used during pretraining. This yields a well-defined separation between \ac{sk} (the induced binary gating patterns) and \ac{qk} (the parameters optimized within those fixed patterns), allowing us to freeze the former and federate only the latter to enable efficient specialization under heterogeneous client data.

\section{Method proposal}
\label{sec:method}
\subsection{Background}
\label{sec:method_b}
In this work, we focus primarily on \acp{cnn}, as they are among the most widely adopted architectures in \ac{dl}. To formalize the setting of our proposal, we first introduce the following two definitions, which provide the necessary background on \ac{cnn} and activation patterns.

\begin{definition}
A \ac{cnn} is a \ac{ff} that maps an input image tensor
$x\in\mathbb{R}^{C_0\times H_0\times W_0}$ to an output $f(x;w)$ by applying a sequence of $L$ convolutional blocks followed by a head. Given $h_0 \coloneqq x,$ we have:
\[
\qquad z_\ell =\mathrm{Conv}_\ell(h_{\ell-1};w_\ell) 
\qquad h_\ell \;=\; \phi_\ell\!\big(z_\ell\big),\quad \ell=1,\dots,L,
\] 
\[
f(x;w) \;=\; g(h_L;w_{\mathrm{out}}).
\]

Here, $\mathrm{Conv}_\ell(\cdot;w_\ell)$ denotes a convolutional layer (with its weights $w_\ell$), $\phi_\ell$ is the activation function (and possibly batch normalization/pooling/residual addition), $z_\ell$ corresponds with the preactivation of layer $\ell$, $h_\ell$ corresponds to the input to layer $\ell + 1$, and $g$ is the final classifier/regressor head. The full parameter set is $w=\{w_1,\dots,w_L,w_{\mathrm{out}}\}$.
\end{definition}

\begin{definition}
Consider a \ac{cnn} with $L$ convolutional blocks that include a pointwise nonlinearity. For an input $x\in\mathbb{R}^{C_0\times H_0\times W_0}$, let
$z_\ell(x;w)$ denote the pre-activation tensor of block $\ell$ (the tensor right before the nonlinearity is applied). We define the \emph{activation mask} of block $\ell$ as
\[
\alpha_\ell(x;w) \;\coloneqq\; \mathds{1}\!\big[z_\ell(x;w)>0\big],
\]
where $\mathds{1}$ indicates the characteristic function, and the inequality is applied elementwise, so $\alpha_\ell(x;w)$ is a binary tensor with the same shape as $z_\ell(x;w)$.
The \emph{activation pattern} induced by $x$ is the collection of all masks across blocks:
\[
\alpha(x;w) \;\coloneqq\; \big(\alpha_1(x;w),\dots,\alpha_L(x;w)\big).
\]
\end{definition}

\subsection{Federated algorithm proposal}
\label{sec:method_a}
In \ac{fl} (\textit{e.g.},~\cite{mcmahan2017,reddi2021}), the goal is to minimize
\[
\mathcal{L}(w) \;=\; \sum_{i=1}^{M} \frac{n_i}{n}\, \mathcal{L}_i(w),
\qquad \text{where } n = \sum_{i=1}^{M} n_i ;
\]
$M$ denotes the number of clients, $n_i$ is the number of samples held by client $i$, and $n$ is the total number of samples across all clients. In addition, the local objective is
\[
\mathcal{L}_i(w) \;=\; \mathbb{E}_{x \sim \mathcal{D}_i}\big[\, \ell_i(w; x)\,\big],
\]
where $\mathcal{D}_i$ is the data distribution of client $i$, and $\ell_i(w; x)$ is the per-sample loss (e.g., cross-entropy) evaluated by the shared parameters $w$ on sample $x$.
We adopt the standard assumptions in~\cite{reddi2021} for nonconvex optimization.

In a standard synchronous federated round $t$, the server samples a subset of clients and broadcasts the current global parameters $w_{t}$. Each selected client $i$ then performs several local optimization steps on its private data, producing either an updated set of parameters $w_{t+1, i}$ or an equivalent update $\Delta w_{t, i}$. The server aggregates these client contributions to obtain the next global iterate $w_{t+1}$, which is then broadcasted to clients for the next round~\cite{mcmahan2017,reddi2021}.

Let $f$ denote the model to be federated, with parameters $w$. Following the DualCopy construction of~\cite{duato2025}, we rewrite the model as a duplicated-parameter approximation with two disjoint copies,
\[
\hat{w} \;=\; \big(w^{\mathrm{SK}},\, w^{\mathrm{QK}}\big).
\]
Intuitively, from a certain point on we duplicate $w$ onto $w^{\mathrm{SK}}$, which encodes the \emph{structural} component that induces (binary) gating regimes; and $w^{\mathrm{QK}}$, which encodes the \emph{quantitative} component of the within-regime mappings. Although this construction doubles the stored parameters, it preserves the number of trainable parameters in our procedure as $w^{\mathrm{SK}}$ is kept frozen while $w^{\mathrm{QK}}$ is optimized and aggregated.

\begin{figure*}[htpb]
    \centering
    \includegraphics[width=\linewidth]{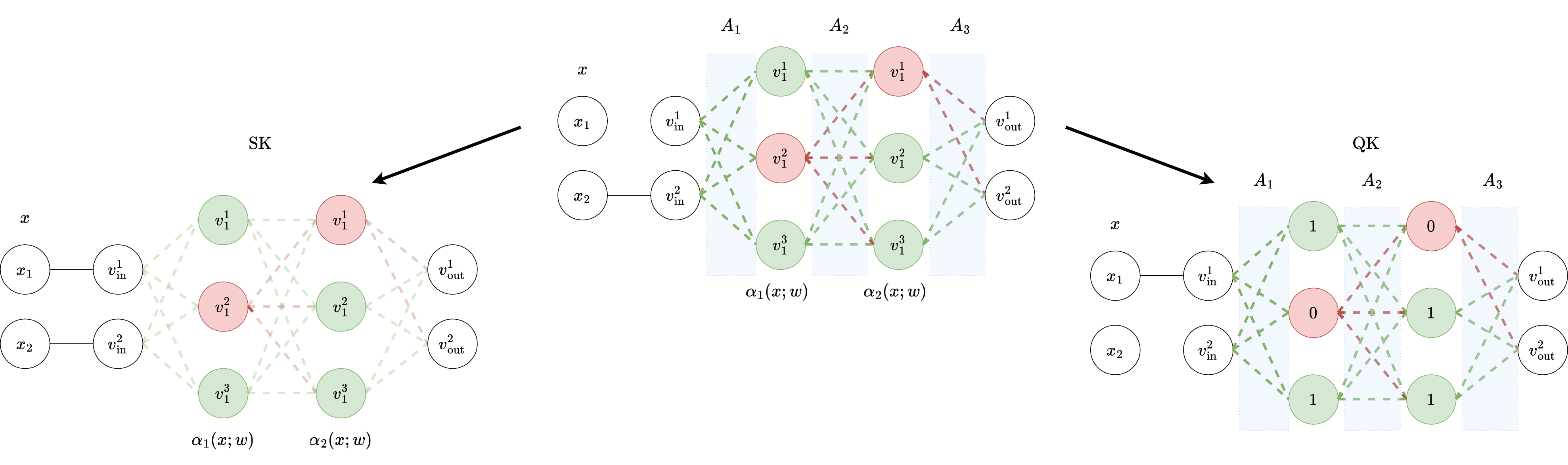}
    \caption{Decoupling view of a network into \ac{sk} representing the values of the activations (left), represented by activation masks (gating); and \ac{qk}, represented by the affine parameters (edge coefficients), with its activation patterns frozen (right). In \textsc{FedSQ}, models are initialized from a common pretrained checkpoint and then adapted federatively by updating only \ac{qk} after structural stabilization.}
    \label{fig:diagram}
\end{figure*}

All participants (server and clients) initialize from a shared pretrained checkpoint $w^{\mathrm{PT}}$. This warm-start is a key departure from training-from-scratch \ac{fl}: the federation begins from a representation that already captures generic visual structure, and subsequent training focuses on efficient downstream adaptation. In our setting, the pretrained initialization instantiates the structural component, i.e., we set $w^{\mathrm{SK}} \leftarrow w^{\mathrm{PT}}$ at initialization. Given an input $x\in\mathbb{R}^{C_0\times H_0\times W_0}$, the structural copy \acs{sk} produces pre-activations
$z_\ell^{\mathrm{SK}}(x)$ and induces a fixed binary mask
$\alpha_\ell(x) \coloneqq \mathds{1}[z_\ell^{\mathrm{SK}}(x) > 0]$.
During the federated stage, the quantitative copy computes
$z_\ell^{\mathrm{QK}}(x)$ and applies the fixed gate:
\[
h_\ell(x) \;=\; \alpha_\ell(x)\odot z_\ell^{\mathrm{QK}}(x),
\]
where $\odot$ corresponds to the elementwise product. Conditioned on $\{\alpha_\ell(x)\}_\ell$, the network reduces to a within-region affine map, where the network behaves as a linear transformation plus bias within the corresponding region~\cite{duato2025}. Hence, freezing $w^{\mathrm{SK}}$ amounts to fixing the per-sample routing mechanism, while federated training updates only the within-region affine parameters encoded by $w^{\mathrm{QK}}$. This knowledge separation is illustrated in Figure~\ref{fig:diagram}.

 Let $\mathcal{D}_0$ denote a calibration set which can be public, synthetic, or obtained via privacy-preserving transformations of client data. Let $S\subset\mathcal{D}_0$ be a fixed probe subset used for evaluation. In this paper, we warm-start from pretrained weights, which can be interpreted as already providing a structurally mature initialization. We propose a centralized phase used to select an effective \ac{tl} strategy for the federation. Specifically, starting from the shared pretrained checkpoint $w^{\mathrm{PT}}$, the server performs a lightweight calibration step on $S$: it does not attempt to train the final federated model, but instead evaluates candidate \ac{tl} schedules (which layers to freeze/unfreeze and in what order) and selects the schedule that yields the most stable and effective adaptation from $w^{\mathrm{PT}}$ on the probe set $S$. The selected schedule is then broadcast to all clients, which follow it during federated training to enforce a consistent adaptation protocol across silos and to mitigate catastrophic forgetting of the pretrained representation.

After this, we freeze the structural component $w^{\mathrm{SK}}$ and run standard synchronous \ac{fl} rounds in which clients update only $w^{\mathrm{QK}}$ on their private data. Concretely, during local training each client uses the frozen $w^{\mathrm{SK}}$ to compute binary gating masks and optimizes $w^{\mathrm{QK}}$ within these fixed activation regimes. Server aggregation is performed on $w^{\mathrm{QK}}$ only, following the same schema as \textsc{FedAvg}.
Algorithm~\ref{alg:FedSQ} summarizes the full procedure. Lines 2-6 correspond to the \ac{tl} regime: the server evaluates candidate \ac{tl} schedules, selects the best-performing one, broadcasts it to all clients, and instantiates the structural component with the pretrained weights. Once $w^{\mathrm{SK}}$ is determined, the server broadcasts it to all clients and the federated stage commences. At each round, the server samples a subset of clients (line 8) and broadcasts the current quantitative parameters. Each selected client then trains locally for a fixed number of epochs on its private data, keeping $w^{\mathrm{SK}}$ frozen and updating only $w^{\mathrm{QK}}$ according to the selected schedule. Finally, the server aggregates the received quantitative parameters (line 15) to form the next global quantitative iterate. This federated loop (lines 7--15) is repeated for $T$ communication rounds.

\begin{algorithm}[htpb]
\caption{\textsc{FedSQ}: transfer-initialized calibration of \ac{tl} schedule + federated \ac{qk} optimization}
\label{alg:FedSQ}
\begin{algorithmic}[1]
  \STATE \textbf{Input:} clients $\{1,\dots,M\}$ with local datasets $\{\mathcal{D}_i\}$ and sizes $\{n_i\}$; $S \in \mathcal{D}_0$ subset;
  pretrained checkpoint $w^{\mathrm{PT}}$; local epochs $E$; total rounds $T$; sampled clients per round $K$.

    \STATE $\sigma \leftarrow \mathrm{ObtainSchedule}\!\left(w^{\mathrm{PT}},\,S\right)$.
    \STATE Broadcast training schedule $\sigma$ to all clients.
    \STATE $w^{\mathrm{SK}} \leftarrow w^{\mathrm{PT}}$.
    \STATE $w^{\mathrm{QK}} \leftarrow w^{\mathrm{PT}}$.

  \STATE Freeze $w^{\mathrm{SK}}$ and broadcast it once to all clients.
  \FOR{$t = 0, \ldots, T-1$}
    \STATE Server samples $K \subseteq M$ clients and broadcasts $w_t^{\mathrm{QK}}$.
    \FOR{each client $i \in K$ \textbf{in parallel}}
      \STATE Client follows schedule $\sigma$ and updates \textbf{only} $w^{\mathrm{QK}}$.
      \STATE Client computes binary gating masks $\{\alpha_\ell(x;w^{\mathrm{SK}})\}_\ell$.
      \STATE Client trains for $E$ local epochs on $\mathcal{D}_i$, obtaining $w_{t+1,i}^{\mathrm{QK}}$.
      \STATE Client sends $w_{t+1,i}^{\mathrm{QK}}$ to the server.
    \ENDFOR
    \STATE Server aggregates QK:
    \[
      w_{t+1}^{\mathrm{QK}} \;=\; \sum_{i\in\mathcal{K}_t}\frac{n_i}{\sum_{j\in\mathcal{K}_t}n_j}\, w_{t+1,i}^{\mathrm{QK}}.
    \]
  \ENDFOR
\end{algorithmic}
\end{algorithm}

\section{Experiments}
The primary objective of this evaluation is to assess the performance of \textsc{FedSQ} against standard federated aggregation schemes under identical experimental conditions, including \textit{(i)} a common pretrained initialization; and \textit{(ii)} a centrally selected \ac{tl} schedule when applicable. Accordingly, our hyperparameter choices are not intended to reflect an optimal or realistic federated tuning procedure, but rather to establish a strong and stable baseline so that performance differences can be attributed to the aggregation strategy itself.

To this end, we consider two representative \acp{cnn}: \textsc{AlexNet}~\cite{alexnet}, due to its simplicity and importance in the \acs{dl} community, and \textsc{ResNet18}~\cite{resnet}, due to its efficiency and wide adoption in computer vision.
\textsc{AlexNet} is trained on CINIC-10~\cite{cinic10}, a proxy based on CIFAR-10 with ImageNet-1K images usage, while \textsc{ResNet18} is trained on CIFAR-100~\cite{cifar100}, a more realistic cross-silo use case.

All experiments were conducted using Python~v3.10, PyTorch~v2.6 and Flower\footnote{https://flower.ai/}~v1.25. To ensure reproducibility, random seeds were fixed across all relevant libraries.  

Hyperparameters were chosen to maximize validation accuracy for the baseline method and then kept fixed across all methods within each model/dataset suite. Concretely, we ran preliminary ablations over a short horizon of $5$ epochs on a grid of learning rate (LR) and weight decay (WD),
$(\mathrm{LR},\mathrm{WD}) \in\{10^{-1},\,10^{-2},\,10^{-3}\}\times\{10^{-3},\,10^{-4}\}$,
and selected $(\mathrm{LR},\mathrm{WD})=(10^{-2},\,10^{-4})$ as it yielded the highest validation accuracy. Unless otherwise stated, all experiments use batch size $64$ per client.
We emphasize that such centralized hyperparameter search may be impractical in fully realistic federated deployments; here it is used solely to ensure a competitive and consistent baseline for comparing aggregation procedures. We also studied results under both \textsc{SGD} and \textsc{AdamW}, using the same selected $(\mathrm{LR},\mathrm{WD})$ pair for fair comparison across aggregation schemes, finally keeping SGD due to better results on the baseline.

\ac{tl} schedules are selected centrally as described in Section~\ref{sec:method_a} using a calibration protocol. In particular, we determine the freezing depth by starting from a conservative policy (freezing up to the penultimate block, i.e., training only the last two blocks) and progressively unfreezing earlier blocks until validation accuracy no longer improves (or starts decreasing), which we interpret as a signal of incipient catastrophic forgetting. Under this criterion, for \textsc{AlexNet} we only train the classifier, whereas for \textsc{ResNet18} we train both the $4^{\text{th}}$ convolutional block and the classifier.

We simulate a cross-silo federation with $M=10$ clients and a single central server. We consider two client partition schemes implemented in the Flower framework: \textit{(i)} an i.i.d.\ split (uniform random assignment of samples); and \textit{(ii)} a heterogeneous Dirichlet split with concentration parameter $\alpha = 0.5$~\cite{yurochkin2019}. We use full participation ($K=M$) and one local epoch per communication round ($E=1$). For \textsc{AlexNet}/CINIC-10 we run $T=5$ rounds since the model saturates quickly (only the classifier is trained under the selected schedule). For \textsc{ResNet18}/CIFAR-100 we run $T=30$ rounds, using as a reference the number of epochs needed to train the model centralized. We compare server-side aggregation strategies under two client-partition regimes. Under an i.i.d.\ partition, we benchmark the standard \textsc{FedAvg} procedure against the proposed \textsc{FedSQ}. Under a heterogeneous Dirichlet partition, we compare \textsc{FedAvg} and \textsc{FedProx} (as it performs better than \textsc{FedAvg} under unbalanced clients) against \textsc{FedSQ}, in order to assess robustness under client drift.

\begin{figure}
    \centering
    \includegraphics[width=\linewidth]{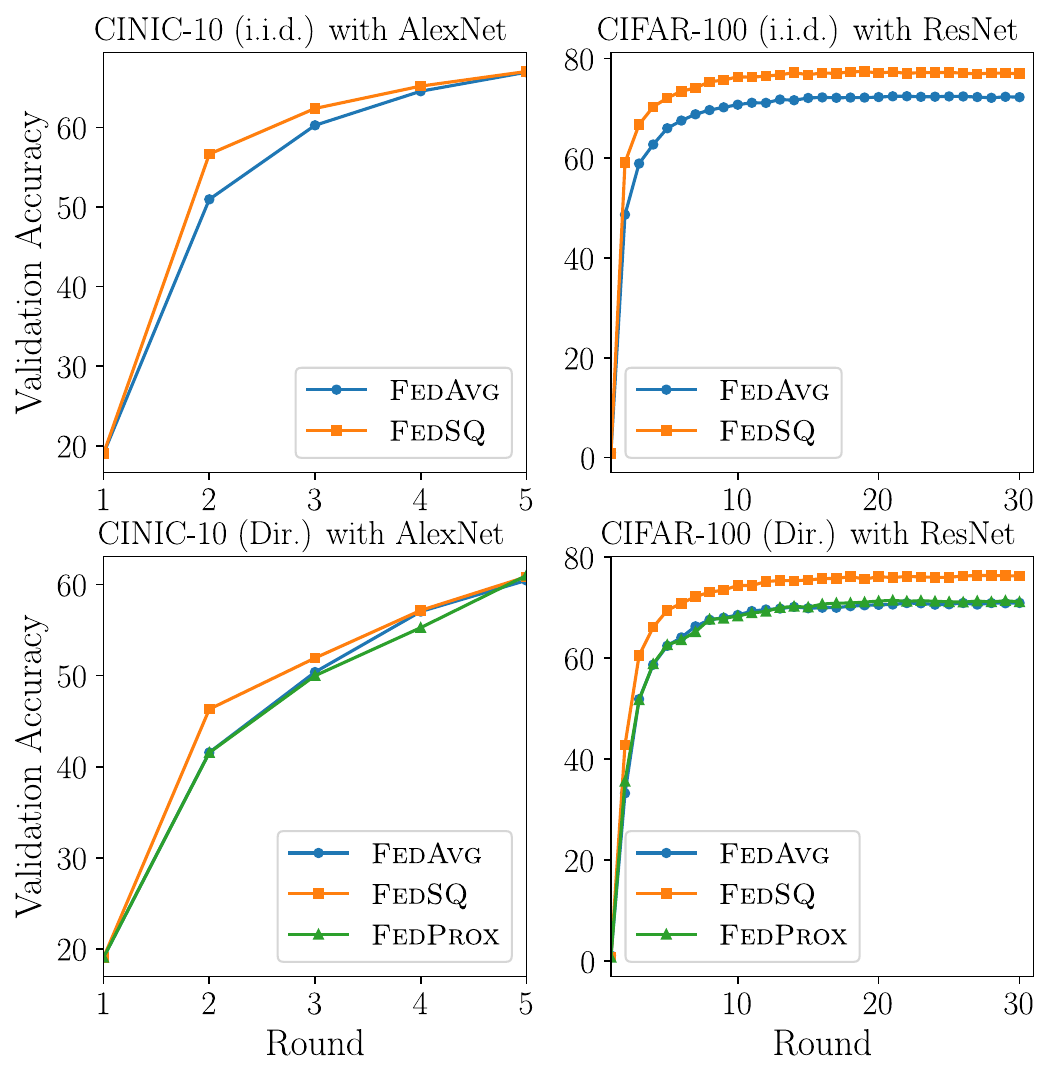}
    \caption{Experiments performed over different model architectures and datasets, showing an improvement of \textsc{FedSQ} over the other approaches. Left column indicates the experiments performed over \textsc{AlexNet}. Right column indicates the experiments performed over \textsc{ResNet18}}
    \label{fig:exps}
\end{figure}
\begin{table}[htpb]
\caption{Comparison of federated algorithms. \ac{bva} denotes the best validation score achieved during training; ``Epoch'' reports the communication round at which \ac{bva} is attained.}
\label{tab:results}
\centering
\begin{tabular}{ccccc}
\toprule
\textbf{Model/Dataset} & \textbf{Part. Scheme} & \textbf{Aggregation} & \textbf{Round} & \textbf{BVA} \\
\midrule
\multirow{5}{*}{\makecell{\textsc{AlexNet}/\\CINIC-10}}
 & \multirow{2}{*}{i.i.d.} 
 & \textsc{FedAvg}   & 5           & \textbf{68.31} \\
 &                   & \textsc{FedSQ} & \textbf{5}  & 68.10 \\
\cmidrule(lr){2-5}
 & \multirow{3}{*}{Dirichlet}
 & \textsc{FedAvg}   & 5           & 62.60 \\
 &                   & \textsc{FedProx}  & \textbf{5}  & 62.30 \\
 &                   & \textsc{FedSQ} & 5           & \textbf{63.00} \\
\midrule
\multirow{5}{*}{\makecell{\textsc{ResNet18}/\\CIFAR-100}}
 & \multirow{2}{*}{i.i.d.}
 & \textsc{FedAvg}   & 21          & 72.44 \\
 &                   & \textsc{FedSQ} & \textbf{18} & \textbf{77.40} \\
\cmidrule(lr){2-5}
 & \multirow{3}{*}{Dirichlet}
 & \textsc{FedAvg}   & 30          & 70.97 \\
 &                   & \textsc{FedProx}  & \textbf{20} & 71.44 \\
 &                   & \textsc{FedSQ} & 30          & \textbf{76.45} \\
\bottomrule
\end{tabular}
\end{table}

Table~\ref{tab:results} summarizes the experimental outcomes. We consider the two client data-partition regimes, both implemented in the Flower framework. Under the i.i.d.\ regime, we compare the standard \textsc{FedAvg} baseline against \textsc{FedSQ}. Under the Dirichlet regime, we additionally include \textsc{FedProx}, as it performs better than \textsc{FedAvg} under unbalanced clients. These results are reported in Table~\ref{tab:results} and visually supported by the learning curves in Figure~\ref{fig:exps}. For \textsc{AlexNet} under i.i.d.\ partitioning, the gap between \textsc{FedAvg} and \textsc{FedSQ} is small (within fractions of a percent), suggesting that when heterogeneity is mild and only the classifier is trained, the benefit of fixed gating is limited, yet \textsc{FedSQ} matches or improves upon the baselines under Dirichlet partitioning, indicating increased robustness when the client distributions diverge. \textsc{ResNet18} on CIFAR-100 exhibits a clear and systematic gain: \textsc{FedSQ} reaches substantially higher best validation accuracy than \textsc{FedAvg} under both i.i.d.\ and Dirichlet splits, and it does so earlier in training. Concretely, \textsc{FedSQ} achieves the best validation score 3 rounds earlier under i.i.d.\ (18 vs.\ 21 rounds) and shows a markedly stronger advantage under Dirichlet heterogeneity (best validation score in the mid-to-late rounds while \textsc{FedAvg} lags behind), as reflected by the separation of curves in Figure~\ref{fig:exps}, which highlights that exploiting the piecewise-affine DualCopy view can make the federated aggregation phase not only more stable, but also more effective in extracting transferable performance under cross-silo heterogeneity. 

\section{Conclusion and future work}
We introduced \textsc{FedSQ}, a federated training procedure motivated by the decoupling between \ac{sk} and \ac{qk}. In \textsc{FedSQ}, a frozen structural component induces binary gating masks, and the federated stage focuses on aggregating only the quantitative copy, effectively learning within-regime affine refinements under a DualCopy piecewise-linear view. Across two \ac{cnn} architectures and heterogeneous client partitions, our results show that this design can improve convergence behavior and robustness under non-i.i.d. splits, and can reach competitive (and in some cases superior) validation accuracy compared to standard baselines.

Beyond the empirical gains, \textsc{FedSQ} provides a practical handle to exploit gating explicitly in \ac{fl}, which has received comparatively less attention than optimizer-centric or personalization-centric approaches. This opens several lines of research. First, once the structural component is fixed, the quantitative updates may admit additional compression without sacrificing performance, for instance via low-rank or sparsified updates, reducing communication costs. Second, the decoupling suggests adaptive system policies: fewer local epochs, fewer participating clients per round, or reduced communication frequency after the structural component has stabilized, trading compute for communication in a controlled manner. Third, while this paper focuses on a transfer-initialized regime, an important next step is to study an alternative initialization pathway in Algorithm~\ref{alg:FedSQ}: learning structural knowledge \emph{from scratch} via a short centralized training phase on a calibration set, freezing it once mature, and then proceeding with federated \ac{qk}-only refinement. Finally, \textsc{FedSQ} appears naturally compatible with asynchronous or partially reliable federated settings, where clients may join intermittently, which lead to a promising line of work, where fixing the gating structure in advance can reduce instability due to drift, making late-stage aggregation easier to stabilize.

\section*{Acknowledgments}
This work was supported by the project C121/23 CIBER-CAFE funded by INCIBE, and by the projects {\small PID2023-146569NB-C21} and {\small PID2023-146569NB-C22} supported by {\small MICIU/AEI/10.13039/501100011033} and ERDF/UE. Alberto Fernández-Hernández was supported by the predoctoral grant {\small PREP2023-001826} supported by {\small MICIU/AEI/10.13039/501100011033} and ESF+. Jose I. Mestre was supported by the project {\small PCI2024-161827-3} funded by {\small MICIU/AEI/10.13039/501100011033} and {\small NextGenerationEU/PRTR}. Manuel F. Dolz was supported by the Plan Gen--T grant {\small CIDEXG/2022/013} of the \emph{Generalitat Valenciana}.

\balance
\bibliographystyle{IEEEtran}

\bibliography{bibliography}
\end{document}